\documentclass{article} 
\usepackage{times}
\usepackage{enumerate}
\usepackage{color}
\usepackage{graphicx,epsfig,subfigure}
\usepackage{algorithm,algorithmic}
\usepackage{amsmath,amssymb,xspace}
\usepackage{url}
\usepackage{subfigure}
\usepackage[hidelinks]{hyperref}

\begin{document}

\newcommand{\var}{{\rm var}}
\newcommand{\Tr}{^{\rm T}}
\newcommand{\vtrans}[2]{{#1}^{(#2)}}
\newcommand{\kron}{\otimes}
\newcommand{\schur}[2]{({#1} | {#2})}
\newcommand{\schurdet}[2]{\left| ({#1} | {#2}) \right|}
\newcommand{\had}{\circ}
\newcommand{\diag}{{\rm diag}}
\newcommand{\invdiag}{\diag^{-1}}
\newcommand{\rank}{{\rm rank}}
\newcommand{\nullsp}{{\rm null}}
\newcommand{\tr}{{\rm tr}}
\renewcommand{\vec}{{\rm vec}}
\newcommand{\vech}{{\rm vech}}
\renewcommand{\det}[1]{\left| #1 \right|}
\newcommand{\pdet}[1]{\left| #1 \right|_{+}}
\newcommand{\pinv}[1]{#1^{+}}
\newcommand{\erf}{{\rm erf}}
\newcommand{\hypergeom}[2]{{}_{#1}F_{#2}}
\newcommand{\mcal}[1]{\mathcal{#1}}

\renewcommand{\a}{{\bf a}}
\renewcommand{\b}{{\bf b}}
\renewcommand{\c}{{\bf c}}
\renewcommand{\d}{{\rm d}}  
\newcommand{\e}{{\bf e}}
\newcommand{\f}{{\bf f}}
\newcommand{\g}{{\bf g}}
\newcommand{\h}{{\bf h}}
\renewcommand{\k}{{\bf k}}
\newcommand{\m}{{\bf m}}
\newcommand{\n}{{\bf n}}
\renewcommand{\o}{{\bf o}}
\newcommand{\p}{{\bf p}}
\newcommand{\q}{{\bf q}}
\renewcommand{\r}{{\bf r}}
\newcommand{\s}{{\bf s}}
\renewcommand{\t}{{\bf t}}
\renewcommand{\u}{{\bf u}}
\renewcommand{\v}{{\bf v}}
\newcommand{\w}{{\bf w}}
\newcommand{\x}{{\bf x}}
\newcommand{\y}{{\bf y}}
\newcommand{\z}{{\bf z}}
\newcommand{\A}{{\bf A}}
\newcommand{\B}{{\bf B}}
\newcommand{\C}{{\bf C}}
\newcommand{\D}{{\bf D}}
\newcommand{\E}{{\bf E}}
\newcommand{\F}{{\bf F}}
\newcommand{\G}{{\bf G}}
\renewcommand{\H}{{\bf H}}
\newcommand{\I}{{\bf I}}
\newcommand{\J}{{\bf J}}
\newcommand{\K}{{\bf K}}
\renewcommand{\L}{{\bf L}}
\newcommand{\M}{{\bf M}}
\newcommand{\N}{\mathcal{N}}  
\renewcommand{\O}{{\bf O}}
\renewcommand{\P}{{\bf P}}
\newcommand{\Q}{{\bf Q}}
\newcommand{\R}{{\bf R}}
\renewcommand{\S}{{\bf S}}
\newcommand{\T}{{\bf T}}
\newcommand{\U}{{\bf U}}
\newcommand{\V}{{\bf V}}
\newcommand{\W}{{\bf W}}
\newcommand{\X}{{\bf X}}
\newcommand{\Y}{{\bf Y}}
\newcommand{\Z}{{\bf Z}}

\newcommand{\bfLambda}{\boldsymbol{\Lambda}}

\newcommand{\bsigma}{\boldsymbol{\sigma}}
\newcommand{\balpha}{\boldsymbol{\alpha}}
\newcommand{\bpsi}{\boldsymbol{\psi}}
\newcommand{\bphi}{\boldsymbol{\phi}}
\newcommand{\bbeta}{\boldsymbol{\beta}}
\newcommand{\Beta}{\boldsymbol{\eta}}
\newcommand{\btau}{\boldsymbol{\tau}}
\newcommand{\bvarphi}{\boldsymbol{\varphi}}
\newcommand{\bzeta}{\boldsymbol{\zeta}}

\newcommand{\blambda}{\boldsymbol{\lambda}}
\newcommand{\bLambda}{\mathbf{\Lambda}}

\newcommand{\btheta}{\boldsymbol{\theta}}
\newcommand{\bpi}{\boldsymbol{\pi}}
\newcommand{\bxi}{\boldsymbol{\xi}}
\newcommand{\bSigma}{\boldsymbol{\Sigma}}
\newcommand{\bPi}{\boldsymbol{\Pi}}
\newcommand{\bOmega}{\boldsymbol{\Omega}}

\newcommand{\bgamma}{\boldsymbol{\gamma}}
\newcommand{\bGamma}{\boldsymbol{\Gamma}}

\newcommand{\bmu}{\boldsymbol{\mu}}
\newcommand{\1}{{\bf 1}}
\newcommand{\0}{{\bf 0}}

\newcommand{\bs}{\backslash}
\newcommand{\ben}{\begin{enumerate}}
\newcommand{\een}{\end{enumerate}}

 \newcommand{\notS}{{\backslash S}}
 \newcommand{\nots}{{\backslash s}}
 \newcommand{\noti}{{\backslash i}}
 \newcommand{\notj}{{\backslash j}}
 \newcommand{\nott}{\backslash t}
 \newcommand{\notone}{{\backslash 1}}
 \newcommand{\nottp}{\backslash t+1}

\newcommand{\notk}{{^{\backslash k}}}
\newcommand{\notij}{{^{\backslash i,j}}}
\newcommand{\notg}{{^{\backslash g}}}
\newcommand{\wnoti}{{_{\w}^{\backslash i}}}
\newcommand{\wnotg}{{_{\w}^{\backslash g}}}
\newcommand{\vnotij}{{_{\v}^{\backslash i,j}}}
\newcommand{\vnotg}{{_{\v}^{\backslash g}}}
\newcommand{\half}{\frac{1}{2}}
\newcommand{\msgb}{m_{t \leftarrow t+1}}
\newcommand{\msgf}{m_{t \rightarrow t+1}}
\newcommand{\msgfp}{m_{t-1 \rightarrow t}}

\newcommand{\proj}[1]{{\rm proj}\negmedspace\left[#1\right]}
\newcommand{\argmin}{\operatornamewithlimits{argmin}}
\newcommand{\argmax}{\operatornamewithlimits{argmax}}

\newcommand{\dif}{\mathrm{d}}
\newcommand{\abs}[1]{\lvert#1\rvert}
\newcommand{\norm}[1]{\lVert#1\rVert}

\newcommand{\eg}{{\em e.g.}\xspace}
\newcommand{\ie}{{i.e.}\xspace}
\newcommand{\SHML}{SHML\xspace}
\newcommand{\ours}{SHML\xspace}
\newcommand{\comment}[1]{}
\newcommand{\sdc}[1]{}
\newcommand{\alanc}[1]{}
\newcommand{\email}[1]{\href{mailto:#1}{#1}}

\title{Supervised Heterogeneous Multiview Learning for Joint Association Study and Disease Diagnosis}
\author{
       Shandian Zhe\\
       Purdue University\\
       \email{szhe@purdue.edu}
       \and
        Zenglin Xu \\
        Purdue University \\
        \email{xu218@purdue.edu}
        \and
        Yuan Qi \\
        Purdue University \\
        \email{alanqi@cs.purdue.edu}
}

\maketitle
\begin{abstract}
Given genetic variations and various phenotypical traits, such as Magnetic Resonance Imaging (MRI) features,  we consider two important and related tasks in biomedical research: i)to select genetic and phenotypical markers for disease diagnosis and ii) to identify associations between genetic and phenotypical data.  These two tasks are tightly coupled because underlying associations between genetic variations and  phenotypical features contain the biological basis for a disease. While a variety of sparse models have been applied for disease diagnosis and canonical correlation analysis and its extensions have bee widely used in association studies (\eg, eQTL analysis), these two tasks have been treated separately. To unify these two tasks, we present a new sparse Bayesian approach for joint association study and disease diagnosis. In this approach, common latent features are extracted from different data sources based on sparse projection matrices and used to predict multiple disease severity levels based on Gaussian process ordinal regression; in return, the disease status is used to guide the discovery of relationships between the data sources. The sparse projection matrices not only reveal interactions between data sources but also select groups of biomarkers related to the disease. To learn the model from data, we develop an efficient variational expectation maximization algorithm. Simulation results demonstrate that our approach achieves higher  accuracy in both predicting ordinal labels and  discovering associations between data sources than alternative methods. We apply our approach to an imaging genetics dataset for the study of Alzheimer's Disease (AD). Our method identifies biologically meaningful relationships between genetic variations, MRI features, and AD status, and achieves significantly higher accuracy for predicting ordinal AD stages than the competing methods. 


\comment{
Given genetic variations and various phenotypical features, such as Magnetic Resonance Imaging (MRI) features,  we consider two related tasks in biomedical research: i)to select genetic and phenotypical markers for disease progression prediction and ii) to identify associations between genetic and phenotypical data.  These two tasks are closely related to each other because underlying associations between genetic variations and  phenotypical features contain the biological basis for disease progression.
While a variety of sparse models have been applied for biomarker selection and canonical-correlation-analysis based association studies have been widely used in various applications (\eg, eQTL analysis), these two tasks have been treated separately. To unify these two tasks,
we present a new sparse Bayesian approach, Supervised Heterogeneous Multiview Learning, for joint association discovery and prediction of disease progression. In this approach, 
common latent features are extracted from different data sources based on sparse projections and used to predict disease status based on Gaussian process ordinal regression; and the information of disease progression  is used to guide the discovery of relationships between the data sources. The sparse projection matrices reveal not only interactions between data sources but also groups of biomarkers related to disease progression.
To learn the model from data, we develop an efficient variational expectation maximization approach that enables us to select the dimension of the latent representation in a principle Bayesian framework. Simulation results demonstrate that our approach achieves higher  accuracy in both predicting disease status and  discovering associations between data sources than alternative methods. Finally, we apply our approach to an imaging genetics dataset for the study of Alzheimer's Disease (AD). Our method identifies promising interactions between
 single nucleotide polymorphisms (SNPs) and MRI features, and achieves significantly higher prediction accuracy for AD progression than the state-of-the-art competing methods.
}



\comment{
Given genetic variations and phenotypical features, a

Alzheimer's Disease (AD) is an irreversible, progressive brain disease that slowly destroys memory and thinking skills and eventually even the basic capability of carrying out daily lives. We study the Alzheimer's Disease based on the Alzheimer's Disease Neuroimaing Initiative (ADNI) data from two perspectives: (1) classifying patients into groups with AD, mild cognitive impairment (MCI), and normal controls, and (2) discovering the associations between the genetic variations (measured on variations of SNPs) and the phenotypic traits (measured on MRI). Although previous research on other biological studies can be borrowed to the analysis on the Alzheimer's Disease,  most of them either study only the association between phenotypic traits and genetic variations neglecting the possible guidance from the patients' status, or focus only on diagnosis neglecting the aid from the association structure.
To address this problem in a principled way, we propose a new approach called Supervised Heterogeneous Multiview Learning (SHML).
Our approach seeks to extract a common latent representation which encodes the common information within all data views, i.e., the genotypes, the phenotypical traits, and the disease progress, aiming to make more accurate diagnosis. Then the latent sparse projections capture the connection between the views,  making it possible to effectively detect the associations with high sensitivity and specificity. In addition, grounded on a Bayesian framework, our method can naturally deal with heterogeneous data types, such as the ordinal values of SNPs, and the continuous values in MRI, and the ordinal values of the disease progress. Experiments on the ADNI data indicates that our approach makes the most accurate diagnosis on the disease progress and finds biologically meaningful associations between SNPs and MRIs.
}

\comment{
In biological and  biomedical research,
the analysis and diagnosis of many complex diseases can be based on a number of data sources or views, such as genetic variations and the phenotypic traits. This brings a new data mining setting where the objectives are of two folds -- to make diagnosis and to study the association between the genetic variations and the phenotypic traits. In this paper, we study the Alzheimer's Disease (AD)

Although these two objectives are highly related, most of previous research  either studies only the association between phenotypic traits and genetic variations neglecting the possible guidance from the patients' status, or focuses only on diagnosis neglecting the aid from the association structure.
To address this problem in a principled way, we propose a new Bayesian framework called Supervised Heterogeneous Multiview Learning (SHML) framework. Our approach seeks to extract a common latent representation which encodes the common information within all data views, i.e., the genotypes, the phenotypical traits, and the healthy status, aiming to make more accurate diagnosis. Then the latent sparse projections capture the connection between the views,  making it possible to effectively detect the associations with high sensitivity and specificity. In addition, grounded on a Bayesian framework, our method can naturally deal with heterogeneous data types, such as the ordinal data type in genetic variations, and the continuous data type in phenotypes. The experiment on a simulated dataset shows that our model achieves significant improvement in both identifying the associations and making prediction when comparing with competitive regression methods and multiview learning methods. Moreover, on the Alzheimer's Disease Neuroimaging Initiative (ADNI) data, our model makes the best diagnosis on the disease level and finds very meaningful associations between SNPs and MRIs.

Alzheimer's Disease (AD) is an irreversible, progressive brain disease that slowly destroys memory and thinking skills and eventually even the basic capability of carrying out daily lives. There is no cure for the disease and it will progress until death. Researchers have collected massive data from AD patients to study the disease mechanism and facilitate the diagnosis. The data can come from different sources or views, such as genetic variations and the phenotypic traits. For example, the Alzheimer's Disease Neuroimaing Initiative (ADNI) data includes both SNPs and MRI imagings of AD patients. It brings a new data mining task where the objectives are of two folds -- to make AD diagnosis and discover the association between the genetic variations (SNPs) and the phenotypic traits (MRI imagings). \sdc{is there some biological proof that shows studying association is a promising direction?} Although these two objectives are highly related, most of previous research either studies only the association between phenotypic traits and genetic variations neglecting the possible guidance from the patients' status, or focuses only on  diagnosis neglecting the aid from the association structure.To address this problem in a principled way, we propose a new Bayesian framework called Supervised Heterogeneous Multiview Learning (SHML) framework. Our approach seeks to extract a common latent representation which encodes the common information within all data views, i.e., the genotypes, the phenotypical traits, and the healthy status, aiming to make more accurate diagnosis. Then the latent sparse projections capture the connection between the views,  making it possible to effectively detect the associations with high sensitivity and specificity. In addition, grounded on a Bayesian framework, our method can naturally deal with heterogeneous data types, such as the ordinal data type in genetic variations (SNPs), and the continuous data type in phenotypes (MRI imagings). Simulation experiments shows that our model achieves significant improvement in both identifying the associations and making prediction when comparing with competitive regression methods and multiview learning methods. On the Alzheimer's Disease Neuroimaging Initiative (ADNI) data, our model makes the best diagnosis on the disease level and finds very meaningful associations between SNPs and MRIs.
}


\end{abstract}

\section{Introduction}

Recent advances in biomedical research have provided
new opportunities to study diseases -- for example, Alzheimer's disease (AD), the most common neurodegenerative disorder -- from multiple data sources. For example, one data source contains genetic variations, such as single nucleotide polymorphisms (SNPs), which can help us understand the genetic basis of diseases. Another data source can be molecular and clinical phenotypes, such as Magnetic Resonance Imaging (MRI) data, which can reveal important phenotypic changes in patients.
Finding associations between different data sources can reveal unknown biological relationships and has a wide range of applications in
computational biology~\cite{Consoli02eqtl}, epidemiology~\cite{Hunter12GWASEpide}, computational neural science~\cite{Gandhi10GWASNeuro}, and imaging genetics~\cite{Liu09snpfmri}.
In addition to  the genotypes and phenotypic traits, we have valuable {\em labeled} information about disease stages from patient medical records. Thus we face a new data analysis setting where the objective is two-fold: i) finding associations between different data sources and ii) selecting relevant (groups of) features from all the sources to predict ordinal disease stages.

Many statistical approaches have been developed to discover associations or select features (or variables) for prediction in a high dimensional problem. For association studies, representative approaches are canonical correlation analysis (CCA) and its extensions~\cite{Hotelling36CCA, Bach05aprobabilistic}. These approaches treat different data sources as separate linear projections from a common latent representation. These approaches have been widely used in expression quantitative trait locus (eQTL) analysis. For example, Parkhomenko et al. \cite{Parkhomenko07scca} applied sparse CCA (sCCA) to find relationships between genetic loci and gene expression levels in Utah families;
Witten and Tibshirani \cite{Witten09scca} used sCCA to reveal associations between gene expression and DNA copy variation;
and Chen et al. \cite{ChenLC12cca} used structured CCA for pathway selection.
For disease diagnosis based on high dimensional biomarkers, popular approaches include
lasso~\cite{Tibshirani94lasso}, elastic net~\cite{Zou05ElasticNet}, and group lasso~\cite{Yuan07groupLasso}, and Bayesian automatic relevance determination~\cite{MacKay91bayesianinterpolation, Neal:1996:BLN:525544}. Here we treat genotypes or phenotypes as predictors (\ie, biomarkers) and the disease status as the response in a linear regression or classification setting. Non-zero estimated regression or classification weights indicate relevant biomarkers for the disease~\cite{Yu12EnrichMCI, Shen10SparseLearningAD}.

Despite their wide success in many applications, these approaches are limited by the following factors:
\begin{itemize}
  \item Most association studies neglect the supervision from the disease status. Because many diseases, such as AD, are a direct result of genetic variations and often highly correlated to clinical traits, the disease status provides useful yet currently unutilized information for finding relationships between genetic variations and clinical traits.
  \item For disease diagnosis, most sparse approaches use classification models and do not consider the order of disease severity. For subjects in AD studies, there is a natural severity order from being normal to 
   mild cognitive impairment (MCI) and then from MCI to AD. Classification models cannot capture the order in AD's severity levels. Furthermore, the classification approaches are often based on conditional models (\eg, logistic regression) and ignore relationships between multiple views.
  \item Most previous methods are not designed to handle heterogeneous data types. The SNPs values are discrete (and ordinal based on an additive genetic model), while the imaging features are continuous.  Popular CCA or lasso-type methods simply treat both of them as continuous data and overlook the heterogeneous nature of the data.
\end{itemize}

To address these problems, we propose a new Bayesian approach that unifies multiview learning with sparse ordinal regression for joint association study and disease diagnosis. In the new approach, genetic variations and phenotypical traits are generated from common {\em latent} features based on separate sparse projection matrices and suitable link functions and the common latent features are used to predict the disease status based on Gaussian process ordinal regression (See Section 2). 
To enforce  sparsity in projection matrices, we assign spike and slab priors \cite{Edward97SpikeSlab} over them; these priors have been shown to be more effective than $l_1$ penalty to learn sparse projection matrices \cite{Goodfellow12s3c, Zoubin12SpikeSlab}. 
The sparse projection matrices not only reveal critical interactions between the different data sources but also identify {\em groups} of biomarkers in data relevant to disease status.
Finding groups of biomarkers can avoid over-sparsification (\ie, selecting one instead of multiple correlated features), thus boosting the accuracy for disease diagnosis. It can also help provide a better biological understanding because these groups may form biologically units (\ie, pathways).
Meanwhile, via its direct connection to the
latent features, the disease status influences the estimation of the projection matrices so that it can guide the discovery of associations  between heterogeneous data sources relevant to the disease.
Hence we name this new method Supervised Heterogeneous Multiview Learning (SHML).

To learn the model from data, we develop a variational Bayesian expectation maximization (VB-EM) approach (See Section 3). It iteratively minimizes the Kullback Leibler divergence between
a tractable approximation and exact Bayesian posterior distributions and provides an estimate to the model marginal likelihood. Maximizing this estimate enables us to automatically choose a suitable dimension for the latent features in a principled Bayesian framework.

In Section 4, we test our approach \ours on both synthetic and real datasets. On synthetic data, \ours achieves both higher estimation accuracy in recovering true associations between different views than CCA and sparse CCA, and higher prediction accuracy than multiple advanced alternative methods, such as the combination of CCA and elastic net, and Gaussian process ordinal regression \cite{ChuGFW05GPOR}.
We then apply \ours to an AD  study. AD
accounts for 60-80\% of age-related dementia cases -- one in eight older Americans has AD -- and there is no cure for AD till now. 
It is believed that its underlying pathology precedes the onset of cognitive symptoms for many years~\cite{Braskie10AD}.
Although AD studies have attracted a lot of attention from both academia and industry \cite{Zhou12FusedLasso,Yuan12mm}, to our best knowledge, our paper presents the first (supervised)  study to uncover associations between genotypes and phenotypic traits relevant to AD.
Our results on Alzheimer's Disease Neuroimaging Initiative (ADNI) data show that \ours achieves highest prediction accuracy among all the competing methods. Furthermore, \ours finds biologically meaningful predictive relationships between SNPs, MRI features, and AD status.

\comment{
All these results indicate that the proposed SHML framework is a new promising tool for analyzing the associations between genotypes and phenotypic traits, and for diagnosis.
}


\section{Model} \label{sec:model}
First, let us describe the data. We assume there are two heterogeneous data sources: one contains continuous data -- for example, MRI features -- and one discrete ordinal data -- for instance, SNPs. Note that we can easily generalize our model below to handle more views and other data types by adopting suitable link functions (\eg, a Possion model for count data). Given data from $n$ subjects, $p$ continuous features and $q$ discrete features, we denote the continuous data by a  $p \times n$ matrix $\X=[\x_1, \ldots, \x_n]$, the discrete ordinal data by a $q \times n$ matrix $\Z=[\z_1, \ldots, \z_n]$, and the labels (\ie, the disease status) by a $n \times 1$ vector $\y=[y_1, \ldots, y_n]^\top$. For the AD study, we let $y_i=0,1,\textrm{ and }2$ if the $i$-th subject is in the normal, MCI or AD condition, respectively.

To link two data sources $\X$ and $\Z$ together, we introduce common latent features $\U=[\u_1, \ldots, \u_n]$
and assume $\X$ and $\Z$ are generated from $\U$ by sparse projection. The common latent feature assumption is sensible for association studies because both SNPs and MRI features are biological measurements of the same subjects. Note that $\u_i$ is the latent feature for the $i$-th subject and its dimension $k$ is estimated by evidence maximization.\alanc{make sure it is there}
In a Bayesian framework, we give a Gaussian prior over $\U$, $p(\U) = \prod_i \N(\u_i|\0, \I)$, and specify the rest of the model (see Figure \ref{fig:graphical_model}) as follows:
\begin{figure}[t!]
\centering
\includegraphics[width=0.4\textwidth]{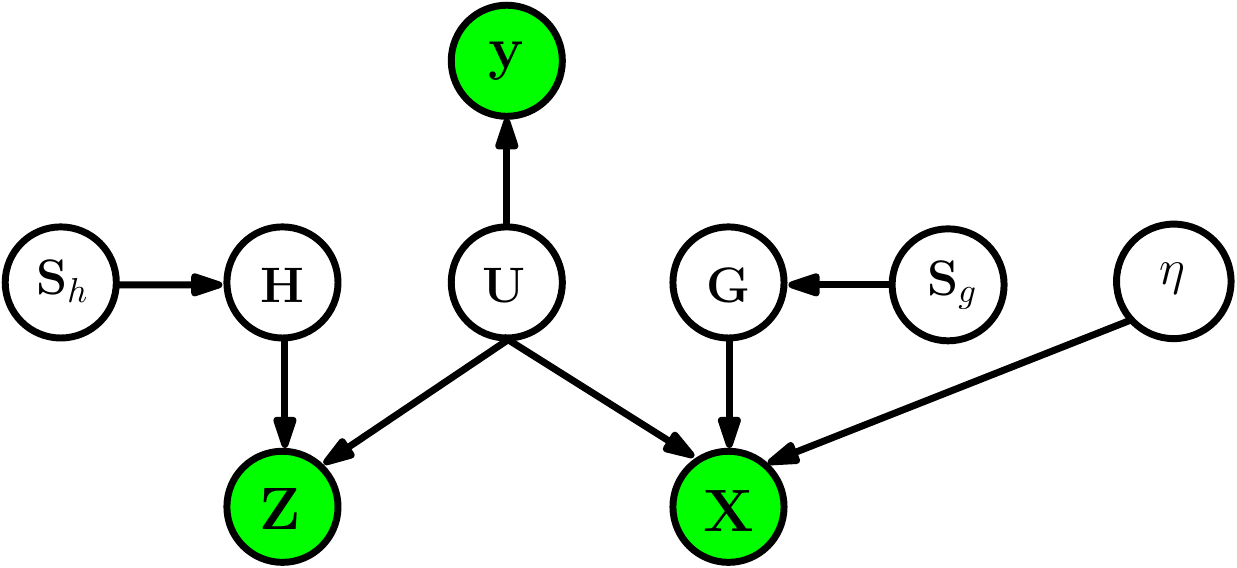}
\caption{\small The graphical model of Supervised Heterogeneous Multiview Learning, where $\X$ is the continuous view, $\Z$ is the ordinal view, and $\y$ are the labels.
 }
\label{fig:graphical_model}
\end{figure}
\begin{itemize}
\item \textbf{Continuous data.} Given $\U$, $\X$ is generated from
\[
p(\X|\U, \G, \eta)= \prod_{i=1}^n \N(\x_i | \G\u_i, \eta^{-1}\I)
\]
where $\G=[\g_1, \g_2, ...\g_p]^\top$ is a $p \times k$ projection matrix, $\I$ is an identity matrix, and $\eta^{-1}\I$ is the precision matrix of the Gaussian distribution. We assign a Gamma prior over $\eta$,
$p(\eta|r_1, r_2) = \textrm{Gamma}(\eta|r_1, r_2)$
        where $r_1$ and $r_2$ are the hyperparameters and set to be $10^{-3}$ in our experiments.

\item \textbf{Ordinal data.} For an ordinal observation $z \in \{0,1,\ldots,$\\$R-1\}$
its value is decided by which region an auxiliary variable $c$ falls in
\[ -\infty=b_0<b_1<\ldots<b_{R}=\infty .\]
If $c$ falls in $[b_{r}, b_{r+1})$, $z$ is set to be $r$.  For the  AD study, the SNPs $\Z$ takes values in  $\{0,1,2\}$ and therefore $R=3$. Given a
$q \times k$ projection matrix $\H=[\h_1, \h_2, ...\h_q]^\top$, the auxiliary variables  $\C=\{c_{ij}\}$ and the ordinal data $\Z$ are generated from
\begin{align}
p(\Z, \C|\U, \H) &= \prod_{i=1}^q\prod_{j=1}^n p(c_{ij}|\h_i, \u_j)p(z_{ij}|c_{ij}) \nonumber
\end{align}
where
\begin{align}
p(c_{ij} | \h_i, \u_j) &= \N(c_{ij} | \h_i^\top\u_j, 1)  \nonumber\\
p(z_{ij} | c_{ij}) &= \sum_{r=0}^{2} \delta(z_{ij} = r)\delta( b_r \le c_{ij} < b_{r+1}),  \nonumber
\end{align}
where $\delta(a)=1$ if $a$ is true and $\delta(a)=0$ otherwise.
\item \textbf{Labels.} The disease statuses $\y$ are ordinal variables too.  To generate $\y$, we use  a Gaussian process ordinal regression model \cite{ChuGFW05GPOR} based the latent representation $\U$,
\[
p(\y|\U) = p(\y|\f) p(\f|\U),
\]
where
\begin{align*}
p(\f|\U) &= \N(\f|\0, \K), \\
p(\y|\f) &= \sum_{r=0}^{2} \delta(y_{i} = r)\delta( b_r \le f_{i} < b_{r+1}),
\end{align*}
where 
$K_{ij} = k(\u_i,\u_j)$ is the cross-covariance between
$\u_i$ and $\u_j$.
We can choose $k$ from a rich family of kernel functions such as linear, polynomial, and Gaussian kernels to model relationships between the labels $\y$ and the latent features $\U$.

Note that the labels $\y$ are linked to the data $\X$ and $\Z$ via the latent features $\U$ and the projection matrices $\H$ and $\G$. Due to the sparsity in $\H$ and $\G$, essentially only a few groups of variables in
 $\X$ and $\Z$ are selected to predict $\y$.  Note that each of group is  linked to a feature in $\U$.

\item \textbf{Sparse Priors.} Because we want to identify a few critical interactions between different data sources,
we use spike and slab prior distributions  \cite{Edward97SpikeSlab} to sparsify the projection matrices $\G$ and $\H$. Specifically, we use
a $p \times k$ matrix $\S_g$ to represent the selection of elements in $\G$:
if $s_{ij}=1$,  $g_{ij}$ is selected and follows a Gaussian prior distribution with variance $\sigma^2_1$; if $s_{ij}=0$,  $g_{ij}$ is not selected and forced to almost zero (\ie, sampled from a Gaussian with a  very small variance $\sigma^2_2$). Specifically,  we have the following prior over $\G$:
  \[
   p(\G|\S_g,\bPi_g) = \prod_{i=1}^p\prod_{j=1}^k p(g_{ij}|s_g^{ij})p(s_g^{ij}|\pi_g^{ij})
  \]
    where
    \begin{align*}
    p(g_{ij}|s_g^{ij}) &= s_g^{ij}\N(g_{ij}|0, \sigma^2_1) + (1-s_g^{ij})\N(g_{ij}|0,\sigma^2_2), \\
        p(s_g^{ij}|\pi_g^{ij}) &= {\pi_g^{ij}}^{s_g^{ij}}(1-\pi_g^{ij})^{1-s_g^{ij}},
    \end{align*}
 where  $\pi_g^{ij}$ in $\bPi_g$ is the probability of $s_g^{ij}=1$, and
   $\sigma^2_1 \gg \sigma^2_2$ (in our experiment, we set $\sigma^2_1 = 1$ and  $\sigma^2_2 = 1o^{-6}$).
 To reflect our uncertainty about $\bPi_g$, we assign a Beta hyperprior distribution:
 \[
  p(\mathbf{\Pi}_g|l_1, l_2) = \prod_{i=1}^p\prod_{j=1}^k \textrm{Beta}(\pi_g^{ij}| l_1, l_2),
 \]
where $l_1$ and $l_2$ are hyperparameters. We set $l_1=l_2=1$ in our experiments.
 Similarly, $\H$ is sampled from
    \[
        p(\H|\S_h, \bPi_h) = \prod_{i=1}^q\prod_{j=1}^k p(h_{ij}|s_h^{ij})p(s_h^{ij}|\pi_h^{ij}),
    \]
    where
\begin{align*}
    p(h_{ij}|s_h^{ij}) &= s_h^{ij}\N(h_{ij}|0, \sigma^2_1) + (1-s_h^{ij})\N(h_{ij}|0,\sigma^2_2), \\
        p(s_h^{ij}|\pi_h^{ij}) &= {\pi_h^{ij}}^{s_h^{ij}}(1-\pi_h^{ij})^{1-s_h^{ij}},
\end{align*}
where $\S_h$ are binary selection variables and
$\pi_h^{ij}$ in $\bPi_h$ is the probability of $s_h^{ij}=1$.
Again, we use a Beta hyperprior distribution:
\[
  p(\mathbf{\Pi}_h|d_1, d_2) = \prod_{i=1}^q\prod_{j=1}^k \textrm{Beta}(\pi_h^{ij}| d_1, d_2),
\]
where $d_1$ and $d_2$ are hyperparameters. We set  $d_1=d_2=1$ in our experiments.

\end{itemize}

Based on all these specifications,  the joint distribution of our model, \ours, is
\begin{align}
&p(\X, \Z, \y, \U, \G, \S_g, \bPi_g, \eta,  \C, \H, \S_h, \bPi_h, \f,  )\notag \\
&=p(\X|\U,\G,\eta)p(\G|S_g)p(S_g|\bPi_g)p(\bPi_g|l_1,l_2)p(\eta|r_1,r_2)\notag \\
&\cdot p(\Z,\C|\U,\H)p(\H|\S_h)p(\S_h|\bPi_h)p(\bPi_h|d_1,d_2)\notag \\
&\cdot p(\y|\f)p(\f|\U)p(\U).
\label{eq:joint_prob}
\end{align}

\section{Estimation}
Given the model specified in the previous section, now we present an efficient, principled method to estimate the latent features
$\U$, the projection matrices  $\H$ and $\G$, the selection indicators
$\S_g$ and $\S_h$, the selection probabilities $\bPi_g$ and $\bPi_h$, the variance $\eta$, the auxiliary variables $\C$ for generating ordinal data $\Z$, and the auxiliary variables $\f$ for generating the labels $\y$. In a Bayesian framework, this estimation task amounts to computing their  posterior distributions.

However, computing the exact posteriors turns out to be infeasible since we cannot calculate the normalization constant of the posteriors based on Equation \eqref{eq:joint_prob}. Thus, we resort to a variational Bayesian Expectation Maximization (VB-EM) approach \cite{Beal03VarEM}. More specifically, in the E step, we approximate the posterior distributions
of $\H,\G,\S_g,\S_h, \bPi_g,$\\$ \bPi_h, \eta, \C$ and $\f$
by a factorized distribution
\[
Q(\H)Q(\G)Q(\S_g)Q(\S_h)Q(\bPi_g)Q(\bPi_h)Q(\eta)Q(\C)Q(\f)
\]
and then use the approximate distributions to compute expectations in the M step to optimize the latent features $\U$.

To obtain the variational approximation, we minimize the
Kullback-Leibler (KL) divergence between the approximate and the exact posteriors,
$KL(Q||P)$ where $P$ represents the exact joint posterior distributions. To this end, we use coordinate descent; we update an approximate distribution, say, $Q(\H)$, while fixing the other approximate distributions, and iteratively refine all the approximate distributions. The detailed updates are given in the following paragraphs.

\subsection{Updating variational distributions for continuous data }\label{sec:con_view}
For the continuous data $\X$, the approximate distributions of the  projection matrix $\G$, the noise variance $\eta$, the selection indicators $\S_g$ and the selection probabilities $\bPi_g$ are
\begin{align}
Q(\G) &= \prod_{i=1}^p \mcal{N}(\g_i; \blambda_i,\bOmega_i), \label{eq:cv_st}\\
Q(\S_g) &= \prod_{i=1}^p\prod_{j=1}^k \beta_{ij}^{s_g^{ij}}(1-\beta_{ij})^{1-s_g^{ij}},\\
Q(\bPi_g) &= \prod_{i=1}^p\prod_{j=1}^k \textrm{Beta}(\pi_g^{ij}| \tilde{l}_1^{ij}, \tilde{l}_2^{ij}),\\
Q(\eta) &= \textrm{Gamma}(\eta| \tilde{r}_1, \tilde{r}_2). \label{eq:cv_ed}
\end{align}
The mean and covariance of $\g_i$ are calculated as follows:
\begin{align*}
\bOmega_i &= \big(\langle \eta \rangle \U\U^\top + \frac{1}{\sigma_1^2}\textrm{diag}(\langle \s_g^i\rangle)+ \frac{1}{\sigma_2^2}\textrm{diag}(\mathbf{1} - \langle \mathbf{s}_g^i \rangle)\big)^{-1},\\
\blambda_i &= \bOmega_i(\langle \eta \rangle \U \tilde{\x}_i),
\end{align*}
where $\langle \cdot \rangle$ means expectation over a distribution,
$\tilde{\x}_i$ and $\s_g^i$ are the transpose of the $i$-th rows of $\X$ and $\S_g$, $\langle \s_g^i\rangle=[\beta_{i1}, \ldots, \beta_{ik}]^\top$, and  $\langle g_{ij}^2\rangle$ is the $j$-th diagonal element in $\bOmega_i$.

The parameter $\beta_{ij}$ in $Q(s_g^{ij})$ is calculated as
$\beta_{ij} = 1/\big(1+\exp(\langle \log(1-\pi_g^{ij}) \rangle - \langle \log(\pi_g^{ij}) \rangle
+ \frac{1}{2}\log(\frac{\sigma_1^2}{\sigma_2^2}) + \frac{1}{2}\langle g_{ij}^2 \rangle(\frac{1}{\sigma_1^2} - \frac{1}{\sigma_2^2}))\big)$.
The parameters of the Beta distribution $Q(\pi_g^{ij})$ is given by $\tilde{l}_1^{ij} = \beta_{ij} + l_1$ and $\tilde{l}_2^{ij} = 1 - \beta_{ij} + l_2$. The parameters of the Gamma distribution $Q(\eta)$ are updated as $\tilde{r}_1 = r_1 + \frac{np}{2}$ and
$\tilde{r}_2 = r_2 + \frac{1}{2}\textrm{tr}(\X\X^\top) - \textrm{tr}(\langle \G \rangle \U \X^\top)
+ \frac{1}{2}\textrm{tr}(\U\U^\top \langle \G^\top\G \rangle)$.


%

The  moments required in the above distributions are calculated as
$\langle \eta \rangle = \frac{\tilde{r}_1}{\tilde{r}_2}$  and
\begin{align}
\langle \log(\pi_g^{ij})\rangle & = \psi(\tilde{l}_1^{ij}) - \psi(\tilde{l}_1^{ij} + \tilde{l}_2^{ij}), \notag \\
\langle \log(1 - \pi_g^{ij})\rangle & = \psi(\tilde{l}_2^{ij}) -
\psi(\tilde{l}_1^{ij} + \tilde{l}_2^{ij}), \notag \\
 \langle \G^\top \G\rangle & = \sum_{i=1}^p \bOmega_i + \blambda_i  \blambda_i^\top, \notag \\
  \langle \G \rangle & =[\blambda_1,\ldots, \blambda_p]^\top,  \label{eq:G}
\end{align}
where $\psi(x)=\frac{\rm{d}}{\rm{dx}}\ln\Gamma(x)$.

\subsection{
Updating variational distributions for ordinal data}\label{sec:ord_view}
For the ordinal data $\Z$, we update the approximate distributions of
the projection matrix $\H$, the auxiliary variables $\C$,  the sparse selection indicators $\S_h$ and the selection probabilities $\bPi_h$. Specifically, the variational distributions of $\C$ and $\H$ are
\begin{align}
Q(\C) &= \prod_{i=1}^q\prod_{j=1}^k Q(c_{ij}),\label{eq:ov_st}\\
Q(c_{ij}) & \propto \delta(b_{z_{ij}} \le c_{ij} < b_{z_{ij}+1})\mcal{N}(c_{ij}| \bar{c}_{ij},1), \label{ordView}\\
Q(\H) &= \prod_{i=1}^q \mcal{N}(\h_i; \bgamma_i,\bLambda_i),
\end{align}
where  $\bar{c}_{ij} = \bgamma_i^\top \u_j$ and
\begin{align*}
\bLambda_i &= \big(\U\U^\top + \frac{1}{\sigma_1^2}\textrm{diag}(\langle \mathbf{s}_h^i \rangle)
+ \frac{1}{\sigma_2^2}\textrm{diag}(\langle \1 - \mathbf{s}_h^i \rangle)\big)^{-1},\\
\bgamma_i &= \bLambda_i(\U \langle \tilde{\mathbf{c}}_i \rangle),
\end{align*}
where $\tilde{\c}_i$ is the transpose of the $i$-th row of $\C$.

The variational distributions of $\S_h$ and  $\bPi_h$ are
\begin{align}
Q(\S_h) &= \prod_{i=1}^q\prod_{j=1}^k \alpha_{ij}^{s_h^{ij}}(1-\alpha_{ij})^{1-s_h^{ij}},\\
Q(\bPi_h) &= \prod_{i=1}^q\prod_{j=1}^k \rm{Beta}(\pi_h^{ij}| \tilde{d}_1^{ij}, \tilde{d}_2^{ij}), \label{eq:ov_ed}
\end{align}
where
$
\alpha_{ij} = 1/\big(1+\exp(\langle \log(1-\pi_h^{ij}) \rangle - \langle \log(\pi_h^{ij}) \rangle
+ \frac{1}{2}\log(\frac{\sigma_1^2}{\sigma_2^2}) + \frac{1}{2}\langle h_{ij}^2 \rangle(\frac{1}{\sigma_1^2} - \frac{1}{\sigma_2^2}))\big) $,
$\tilde{d}_1^{ij} = \alpha_{ij} + d_1$, $\tilde{d}_2^{ij} = 1 - \alpha_{ij} + d_2
$,
 $\langle \s_h^i\rangle =[\alpha_{i1},\ldots,\alpha_{ik}]^\top$, and $\langle h_{ij}^2\rangle$ is the $j$-th diagonal element in $\bLambda_i$.

The required moments for updating the above distributions can be calculated as follows:
\begin{align*}
\langle \tilde{c}_i \rangle  &= [\langle c_{i1} \rangle, \ldots, \langle c_{in}\rangle]^\top,\\
 \langle \log(\pi_h^{ij})\rangle  &= \psi(\tilde{d}_1^{ij}) - \psi(\tilde{d}_1^{ij} + \tilde{d}_2^{ij}), \\
\langle \log(1 - \pi_h^{ij})\rangle & = \psi(\tilde{d}_2^{ij}) - \psi(\tilde{d}_1^{ij} + \tilde{d}_2^{ij}), \\
 \langle c_{ij} \rangle & = \bar{c}_{ij} - \frac{\N(b_{z_{ij}+1}|\bar{c}_{ij},1) - \N(b_{z_{ij}}|\bar{c}_{ij},1)}{\Phi(b_{z_{ij}+1}-\bar{c}_{ij}) - \Phi(b_{z_{ij}} - \bar{c}_{ij})},
\end{align*}
where $\Phi(\cdot)$ is the cumulative distribution function of a standard Gaussian distribution. Note that in Equation~\eqref{ordView}, $Q(c_{ij})$ is a truncated Gaussian and the truncation is controlled by the observed ordinal data $z_{ij}$.


\subsection{Updating variational distributions for labels}
We update the variational distribution of the auxiliary variables $\f$ as follows:
\begin{align}
Q(\f) &= \prod_{i=1}^n Q(f_i) \label{eq:ol_st} ,\\
Q(f_i) &\propto  \delta(b_{y_i} \le f_i < b_{y_i+1}) \mcal{N}(f_i|\bar{f}_i,\sigma_{f_i}^2), \label{eq:ol}
\end{align}
where
\begin{align}
\bar{f}_{i} &= \K_{i,\neg i} \K_{\neg i, \neg i}^{-1} \langle \f_{\neg i} \rangle, \\
\sigma_{f_i}^2 &= \K_{i,i} - \K_{i,\neg i}\K_{\neg i, \neg i}^{-1}\K_{\neg i,i},
\end{align}
where $\K_{i,\neg i}$ is the covariance between $\u_i$ and $\U_{\neg i}$, $\K_{\neg i, \neg i}$
is the covariance on $\U_{\neg i}$
($\U_{\neg i} = [\u_1,\cdots \u_{i-1}, \u_{i+1},\cdots \u_{n}]$), $\langle \f_{\neg i} \rangle = [\langle f_1 \rangle,\cdots, \langle f_{i-1}\rangle, \langle f_{i+1} \rangle,\cdots, \langle f_n \rangle]^\top$, and each $\langle f_i \rangle$ is
\begin{align}
 \langle f_i \rangle &= \bar{f}_i - \sigma_{f_i}^2\cdot \frac{\N(b_{y_i+1}|\bar{f}_i, \sigma_{f_i}^2) - \N(b_{y_i}|\bar{f}_i, \sigma_{f_i}^2)}{\Phi(\frac{b_{y_i+1}-\bar{f}_i}{\sigma_{f_i}}) - \Phi(\frac{b_{y_i} - \bar{f}_i}{\sigma_{f_i}})} \label{eq:ef}.
\end{align}
Note that $Q(f_i)$ is also a truncated Gaussian and the truncated region is decided by the ordinal label $y_i$. 
In this way, the supervised information from $\y$ is incorporated into estimation  of $\f$ and then estimation of the other quantities by the recursive updates.

\subsection{Optimizing the latent representation $\U$}\label{sec:optU}
After the expectations of the other variables are calculated, we optimize $\U$ by maximizing the following variational lower bound
\begin{align}\label{eq:obj_U}
 F(\U) &= -\frac{1}{2}\rm{tr}(\U\U^\top) + \langle \eta \rangle\rm{tr}(X^\top \langle\G \rangle \U) \notag \\
 &- \frac{1}{2}\rm{tr}(\langle \H^\top \H \rangle \U\U^\top) - \frac{1}{2}\rm{log}|\K| - \frac{1}{2}\rm{tr}(\langle \f\f^\top \rangle \K^{-1})\notag \\
&- \frac{\langle \eta \rangle}{2}\rm{tr}(\langle \G^\top\G \rangle\U\U^\top)+ \rm{tr}(\langle \C \rangle^\top \langle \H \rangle\U) + \textrm{constant}, \raisetag{0.36in}
\end{align}
where
\begin{align}
\langle \H^\top \H\rangle &= \sum_{i=1}^p \bLambda_i + \bgamma_i \bgamma_i^\top, \quad \quad \langle \H \rangle = [\h_1, \ldots, \h_q]^\top,\\
\langle\f\f^\top \rangle &= \langle \f  \rangle \langle \f \rangle^\top - \rm{diag}(\langle \f \rangle^2) + \rm{diag}(\langle \f^2\rangle),\\
\langle f_i^2 \rangle & = \langle f_i \rangle^2 + \sigma_{f_i}^2 \notag \\
& \;
+\sigma_{f_i}^2 \cdot \frac{(b_{y_i} - \langle f_i \rangle)\N(b_{y_i}|\langle f_i \rangle, \sigma_{f_i}^2)}{\Phi(\frac{b_{y_i+1} - \langle f_i \rangle}{\sigma_{f_i}}) - \Phi(\frac{b_{y_i} - \langle f_i \rangle}{\sigma_{f_i}})}\notag \\
&\; - \sigma_{f_i}^2 \cdot \frac{(b_{y_i+1} - \langle f_i \rangle)\N(b_{y_i+1}|\langle f_i \rangle, \sigma_{f_i}^2)}{\Phi(\frac{b_{y_i+1} - \langle f_i \rangle}{\sigma_{f_i}}) - \Phi(\frac{b_{y_i} - \langle f_i \rangle}{\sigma_{f_i}})},
\end{align}
and the constant means a value independent of $\U$ so that it is irrelevant for optimizing $\U$.  Note that we can optimize the dimension $k$ by maximizing the full variational lower bound of our model, which involves other quantities as well, such as $\langle \H \rangle$ and  $\langle \G \rangle$. To save space, we do not present the long equation for the full lower bound (which can be easily derived based on what we have presented).

The other required moments are given in Equations \eqref{eq:G} and \eqref{eq:ef}. We use the L-BFGS algorithm to maximize the cost function $F$ over  $\U$. The gradient of $\U$ is given by
\begin{align}\label{eq:gradient_U}
\frac{\partial F}{\partial \U} &= \langle \eta \rangle \langle \G \rangle^\top \X + \langle \H \rangle^\top \langle \C \rangle
-\big(\I + \langle \eta \rangle \langle \G^\top\G \rangle \notag \\
& + \langle \H^\top\H \rangle\big) \U -\frac{1}{2}\big(\K^{-1} - \frac{1}{2}\K^{-1}\langle \f\f^\top \rangle \K^{-1}\big)\frac{\partial \K}{\partial \U}.
\end{align}
Note that $\frac{\partial \K}{\partial \U}$ depends on the form of the  kernel function $k(\u_i, \u_j)$.

\begin{algorithm}[tbh!]
\centering
\caption{VB-EM for model estimation}
\begin{tabular*}{3in}{l}
\textbf{1.} Initialize $\U$, the hyperparamters, and the \\
\hspace{5.mm}moments of all the approximate distributions. \\
\textbf{2.} Loop until convergence:\\ 
\hspace{7.mm} \textbf{E-step} Update all the approximate dis-\\
\hspace{7.mm} tributions according to (\ref{eq:cv_st}-\ref{eq:cv_ed}, \ref{eq:ov_st}-\ref{eq:ov_ed}, \ref{eq:ol_st}-\ref{eq:ol}). \\
\hspace{7.mm} \textbf{M-step} Use L-BFGS to optimize $\U$. \\
\textbf{3.} Output $\U$ and all the approximate posterior\\
\hspace{4.mm} distributions.
\end{tabular*}
\\
\vspace{.05in}
\label{alg:vara}
\end{algorithm}

\subsection{Prediction}
Let us denote the training data as $\mcal{D}_{\rm{train}}=\{\X_{\rm{train}},\Z_{\rm{train}},$\\$\y_{\rm{train}} \}$ and the test data as $\mcal{D}_{\rm{test}}=\{\X_{\rm{test}}, \Z_{\rm{test}}\}$. The prediction task needs the latent representation $\U_{\rm{test}}$ for $\mcal{D}_{\rm{test}}$. There are two candidate strategies for obtaining $\U_{\rm{test}}$. The first one is separate learning: we first learn the projection matrices from $\mcal{D}_{\rm{train}}$, i.e., $Q(\H)$ and $Q(\G)$, and then fix them in the variational EM procedure on $\mcal{D}_{\rm{test}}$ to learn $\U_{\rm{test}}$. Note that there are no updates for ordinal label part on $\D_{\rm{test}}$ and the terms regarding ordinal labels should also be removed from Equation \eqref{eq:obj_U} and \eqref{eq:gradient_U}. The second strategy  is joint learning, where we carry out variational EM simultaneously on $\mcal{D}_{\rm{train}}$ and $\mcal{D}_{\rm{test}}$. A drawback of the first strategy is that the (distributions of) loading matrices are fixed when learning latent representation $\U_{\rm{test}}$. Therefore, we adopt the the second strategy; in other words, the variational EM algorithm uses all the data to update the variation distributions, except $Q(\f)$ where only labels in the training set are used. After both $\U_{\rm{test}}$ and $\U_{\rm{train}}$ are obtained from the M-step, we predict the labels for test data as follows:
\begin{align}
\f_{\rm{test}} &= \K\big(\U_{\rm{test}}, \U_{\rm{train}}\big)\K^{-1}\big(\U_{\rm{train}},\U_{\rm{train}}\big)\langle \f_{\rm{train}} \rangle ,\\
y_{\rm{test}}^i &= \sum_{r=0}^{R-1} r\cdot \delta(b_{r} \le f_{\rm{test}}^i < b_{r+1}),
\end{align}
where $y_{\rm{test}}^i$ is the prediction for $i$-th test sample.

\section{Related Work}
\label{sec:related}
The proposed \ours model is related to a broad family of probabilistic latent variable models, including probabilistic principle component analysis~\cite{Tipping99PPCA}, probabilistic canonical correlation analysis~\cite{Bach05pcca} and their extensions~\cite{GuanD09sparseppca,Yu06supervisedprobabilistic,Archambeau08spp,Virtanen11bcca}.They all learn a latent representation whose projection leads to the observed data. Recent studies on probabilistic factor analysis methods put more focus on the sparsity-inducing priors to the projection matrix. Among them, Guan {\em et al.}~\cite{GuanD09sparseppca} used the Laplace prior, the Jeffrey's prior, and the inverse-Gaussian prior; Archambeau \& Bach~\cite{Archambeau08spp} employed the inverse-Gamma prior; and Virtanen {\em et al.} ~\cite{Virtanen11bcca} used the Automatic Relevance Determination(ARD) prior. Despite their success, these sparsity-inducing priors have their own disadvantages -- they confound the degree of sparsity with the degree of regularization on both relevant and irrelevant variables, while in practical settings there is little reason that these two types of complexity control should be so tightly bounded together. Although the inverse-Gaussian prior and the inverse-Gamma prior provide more flexibility of controlling the sparsity, they suffer from being highly sensitive to the controlling parameters and thus lead to unstable solutions. In contrast, our model adopts the spike and slab prior, which has been recently used in multi-task multiple kernel learning~\cite{TitsiasL11SSMTMKL}, sparse coding~\cite{Goodfellow12s3c}, and latent factor analysis~\cite{Carvalho08ss}.
Note that while our Beta priors over the selection indicators lead to simple yet effective variational updates, the hierarchical prior in \cite{Carvalho08ss} can  better handle the selection uncertainty.
Regardless what priors are assigned to the spike and slab models, they generally avoid the confounding issue by separately controlling  the projection sparsity and the regularization effect over selected elements.

SHML is also connected with many methods on learning from multiple sources or views~\cite{Hardoon08workshop}. Multiview learning methods are often used to learn a better classifier for multi-label classification -- usually in text mining and image classification domains -- based on correlation structures among the training data and the labels~\cite{Yu06supervisedprobabilistic,Virtanen11bcca,Rish08sdr}. However, in medical analysis and diagnosis, we meet two separate tasks -- the association discovery between genetic variations and  clinical traits, and the diagnosis on patients. Our proposed SHML conducts these two tasks simultaneously: it employs the diagnosis labels to guide  association discovery, while leveraging the association structures to improve the diagnosis. In particular, the diagnosis procedure in SHML leads to an ordinal regression model based on latent Gaussian process models.
The latent Gaussian process treatment differentiates ours from multiview CCA models \cite{Rupnik10mvcca}. Moreover, most multiview learning methods do not model the heterogeneous data types from different views, and simply treat them as continuous data. This simplification can  degrate the predictive performance. Instead, based on a probabilistic framework , SHML uses suitable link functions to fit different types of data.


\section{Experiments}
\label{sec:exp}
In this section, we demonstrate the effectiveness of \ours on both synthetic and real data for AD study.

\subsection{Simulation Study}

We first design a simulation study to
examine the performance of \ours in terms of (i) estimation accuracy in finding  associations  between two views and (ii) prediction accuracy on ordinal labels.

\textbf{Simulation data}.
To generate the ground truth, we set $n=200$ (200 instances),
$p=q=40$, and  $k=5$.
We designed $\G$, the $40 \times 5$ projection matrix for  the continuous  data $\X$, to be a block diagonal matrix; each column of $\G$ had $8$ elements being ones and the rest of them were zeros, ensuring each row with only one nonzero element.
We designed $\H$, the $40 \times 5$ projection matrix for  the ordinal data $\Z$, to be a block diagonal matrix; each of the first four columns of $\H$ had $10$ elements being ones and the rest of them were zeros, and the fifth column contains only zeros. We randomly generated the  latent representations $\U\in \mathbb{R}^{k\times n}$ with each column $\u_i\sim \mcal{N}(\0, \I)$.
To generate $\Z$, we first sampled the auxiliary variables $\C$ with each column $\c_{i} \sim \N(\H\u_i, 1)$, and then decided the value of each element $z_{ij}$ in $\Z$ by the region $c_{ij}$ falls in -- in other words, $z_{ij}  = \sum_{r=0}^{2}  r \delta( b_r<c_{ij} \le b_{r+1})$ where $\b=\{-\inf, -1, 1, \inf\}$. Similarly, to generate $\y$, we sampled the auxiliary variables $\f$ from $\N(0, \U^\top\U+ \I)$ and then each $y_i$ was generated by $p(y_i | f_i) = \delta(y_i = 0)\delta(f_i \le 0) + \delta(y_i=1)\delta(f_i > 0)$.

\textbf{Comparative methods}. We compared \ours with several state-of-the-art methods including (1) CCA \cite{Bach05aprobabilistic}, which finds the projection directions that maximize the correlation between two views, (2) sparse CCA \cite{Sun11LSCCA,Witten09scca}, where sparse priors are put on the CCA directions, and (3) Multiple Regression with lasso (MRLasso) \cite{Kim09mulreg} where each column of the second view ($\Z$) is regarded as the output of the first view ($\X$).
We did not include results from the sparse probabilistic projection approach \cite{Archambeau08spp} because it performed unstably in our experiments.  Regarding the software implementation,
we used the built-in Matlab Matlab routine for CCA and the code by
\cite{Sun11LSCCA}  for sparse CCA. We implemented MRLasso based on the Glmnet package (\url{cran.r-project.org/web/packages/glmnet/index.html}).

To compare accuracy on predicting labels $\y$, we compared our method with the following ordinal or multinomial regression methods: (1) lasso for multinomial regression \cite{Tibshirani94lasso}, (2) elastic net for multinomial regression \cite{Zou05ElasticNet}, (3) sparse ordinal regression with the splike and slab prior, (4) CCA + lasso, for which we first ran CCA to obtain the latent features $\H$ and then applied lasso to predict $\y$, (5) CCA + elastic net, for which we first ran CCA to obtain the projection matrices and then applied elastic net on the projected data, (6) Gaussian Process Ordinal Regression (GPOR) \cite{ChuGFW05GPOR}, which employs Gaussian processes to learn the latent function for ordinal regression, and (7) Laplacian Support Vector Machine (LapSVM) \cite{melacci2011primallapsvm}, a semi-supervised SVM classification method. We used the Glmnet package for lasso and elastic net, the GPOR package by \cite{ChuGFW05GPOR}, and the LapSVM package by \cite{melacci2011primallapsvm}. For all the methods, we used 10-fold cross validation to tune free parameters for each run; for example, we used extensive cross-validation to choose the kernel form (Gaussian or Polynomials) and its parameters (the kernel width or polynomial orders) for \ours, GPOR, and LapSVM.
Note that all these methods, except \ours, stack $\X$ and $\Z$ together into one data matrix and ignore their heterogeneous nature.

Because alternative methods cannot learn the dimension automatically from the data, for fair comparison, we provided the dimension of the latent representation to all the methods we tested in our simulations.
For each run in our experiment, we partitioned the data into 10 subsets and used 9 of them for training and 1 subset for testing. We repeated the procedure 10 times to generate the averaged results.

\begin{figure}[ht]
\centering
\includegraphics[scale=0.5]{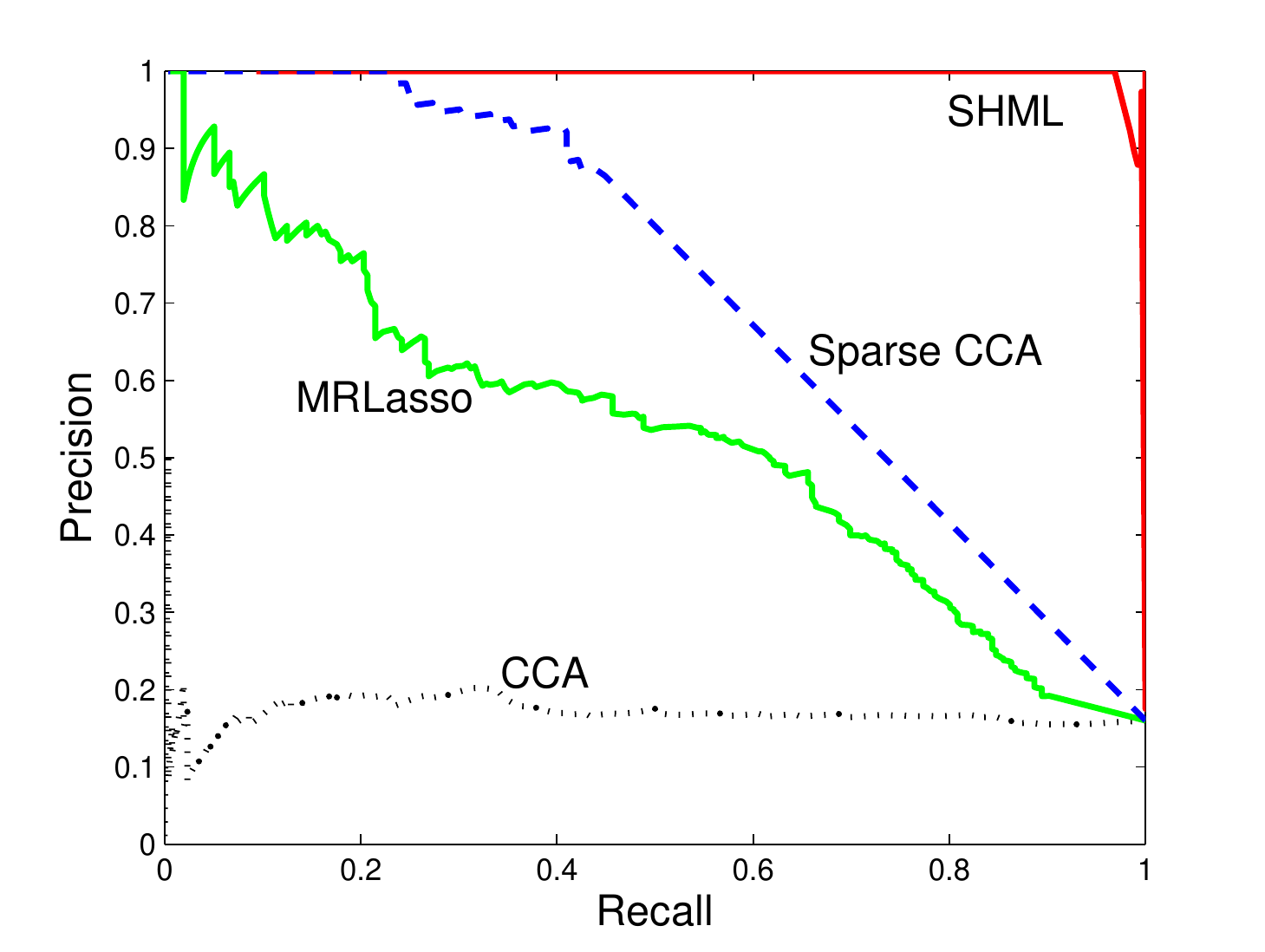}
\caption{\small The precision-recall curves for association discovery.}
\label{fig:toy_roc}
\vspace{-5pt}
\end{figure}

\textbf{Results}.
To estimate linkage (\ie, interactions) between $\X$ and $\Z$, we calculated the cross covariance matrix $\G\H^\top$. We then
computed the precision and the recall based on the ground truth. The
 the precision-recall curves are shown in Figure \ref{fig:toy_roc}. Clearly, our method successfully recovered almost all the links and significantly outperformed all the competing methods. This improvement may come from i) the use of the spike and slab priors, which not only remove irrelevant elements in the projection matrices but also avoid over-penalize the active association structures (the Laplace prior used in sparse CCA does over penalize the relevant ones) \alanc{add to model.tex} and ii) more importantly, the supervision from the labels $\y$, which is probably the biggest difference between ours and the other methods for the association study.

\begin{figure}[ht]
\vspace{-0.1in}
\centering
\includegraphics[scale=0.45]{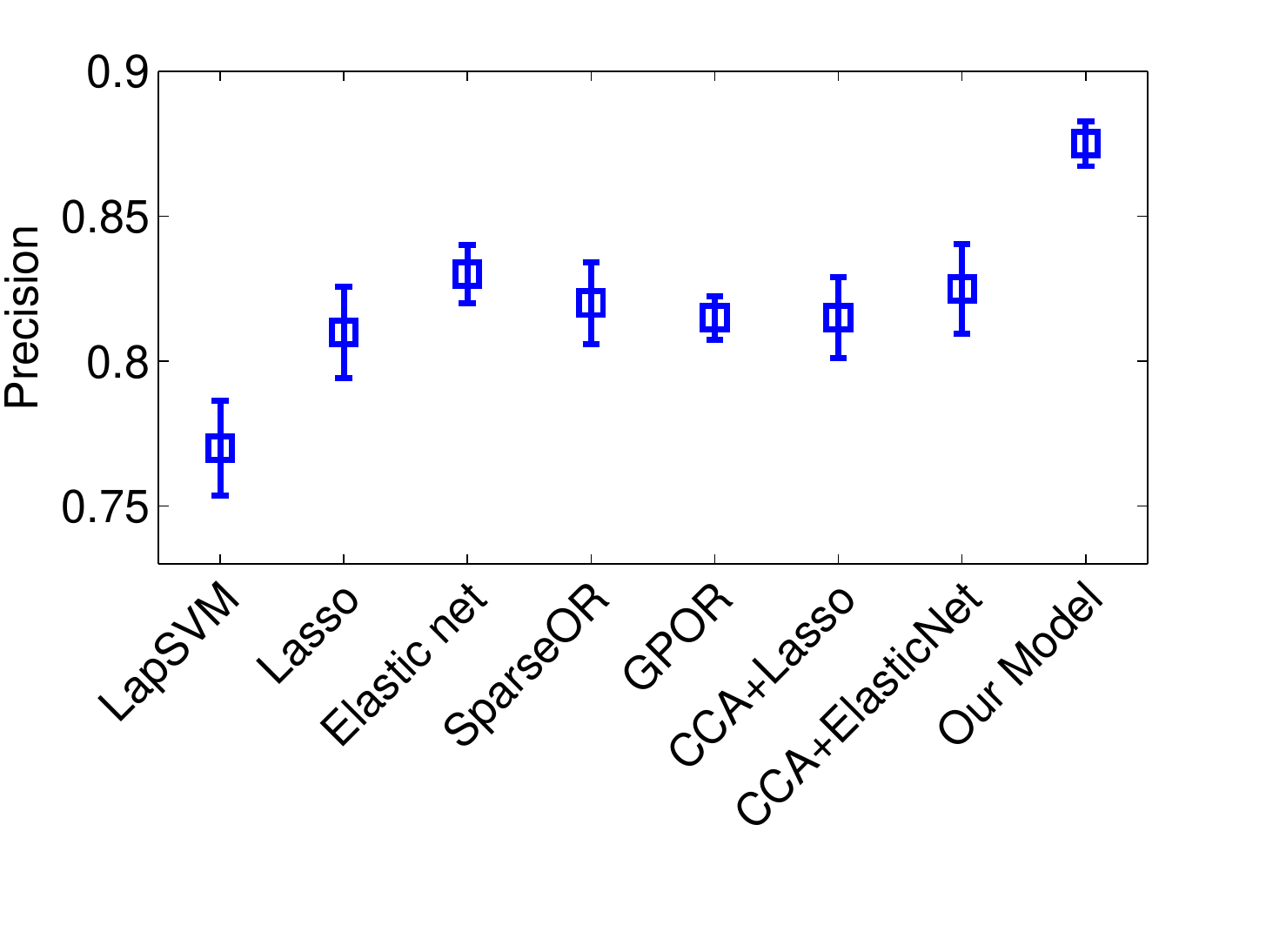}
\vspace{-25pt}
\caption{\small Prediction accuracies on the simulation data. The results are averaged over 10 runs.
 }
\vspace{-5pt}
\label{fig:toy_acc}
\end{figure}

The prediction accuracies on unknown $\y$ and their standard errors are shown in Figure \ref{fig:toy_acc}.
Our proposed \SHML model achieves significant improvement over all the other methods.
\comment{
 Finally, we report the average prediction accuracy with standard errors on the test data in Figure \ref{fig:toy_acc}. It can be found that our proposed method \SHML can achieve significant improvement over all the other methods in terms of prediction accuracy.}
In particular, it reduces the prediction error of elastic net (which ranks the second best) by 25\%, and reduces the error of LapSVM (which ranks the last), by 48\%.
Note that although utilizing the information from the unlabeled data, LapSMV lacks the capability to utilize underlying interaction structures and sparsify the model parameters, which may contribute its poor performance in the experiments.

In summary, the simulation results confirm the power of \ours in both discovering true associations between heterogeneous data sources and predicting unknown labels.

\subsection{Study of Alzheimer's Disease}
We conducted association analysis and diagnosis of AD based on a dataset from Alzheimer's Disease Neuroimaging Initiative(ADNI). The ADNI study is a longitudinal multisite observational study of elderly individuals with normal cognition, mild cognitive impairment, or AD. AD is the most common form of dementia with about 30 million patients worldwide and
payments for care are estimated to be \$200 billion in 2012.\footnote{\url{www.alz.org/downloads/facts_figures_2012.pdf}}.
 In this analysis, we used SHML to study the associations of genotypes and brain atrophy measured by MRI and to predict the subject status (normal vs MCI vs AD).  Note that the labels are ordinal since the three states represent increasing severity levels of the dementia.

The dataset was downloaded from \url{http://adni.loni.ucla.edu/}. After removing missing data, it consists 618 subjects (183 normal, 308 MCI and 134 AD), and for each patient, there are 924 SNPs (selected as the top SNPs to separate normal subjects from AD in ADNI) and 328 MRI features measuring the brain atrophies in different brain regions based on cortical thickness, surface area or volume using FreeSurfer software.

\begin{figure}[ht]
\centering
\includegraphics[scale=0.4]{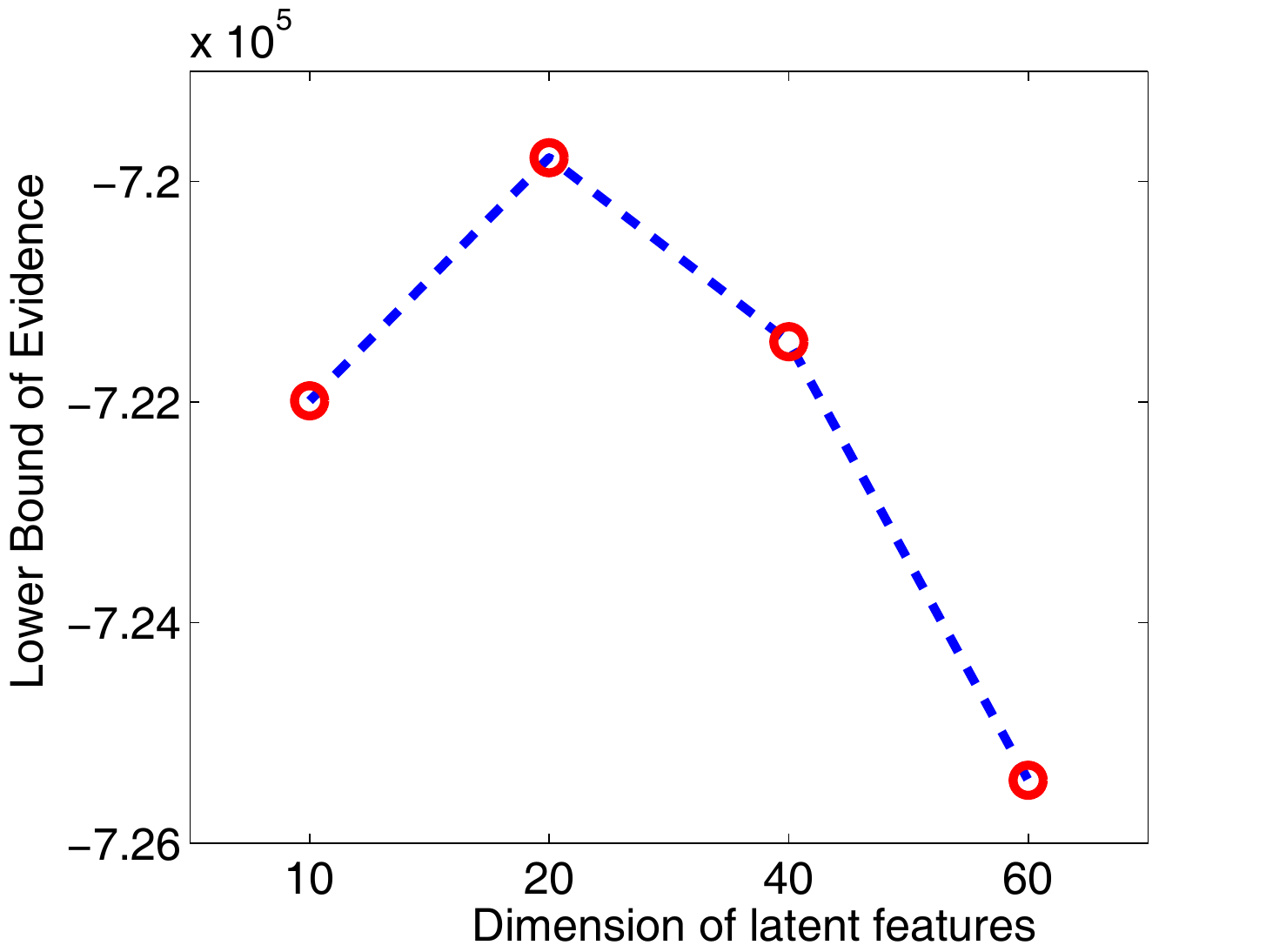}
\caption{\small The variational lower bound for the model marginal likelihood.
 }
\label{fig:real_evidence}
\end{figure}

We compared \ours with the alternative methods on accuracy of predicting whether a subject is in the normal or MCI or AD condition.
We randomly split the dataset into 556 training and 62
test samples 10 times and ran all the competing methods on
each partition.
 As for the simulation study, we used the 10-fold cross validation for each run to tune free parameters.
 In \SHML, in order to determine dimension $k$ for the latent representation $\U$, we computed the variational lower bounds as an approximation to the model marginal likelihood (\ie, evidence), 
 with  various $k$ values $\{10, 20, 40, 60\}$. We chose the value with the largest approximate evidence, which led to $k=20$ (see Figure~\ref{fig:real_evidence}).
Our experiments confirmed that with $k=20$, \ours achieved highest prediction accuracy, demonstrating the benefit of evidence maximization.

\begin{figure}[ht]
\centering
\includegraphics[scale=0.45]{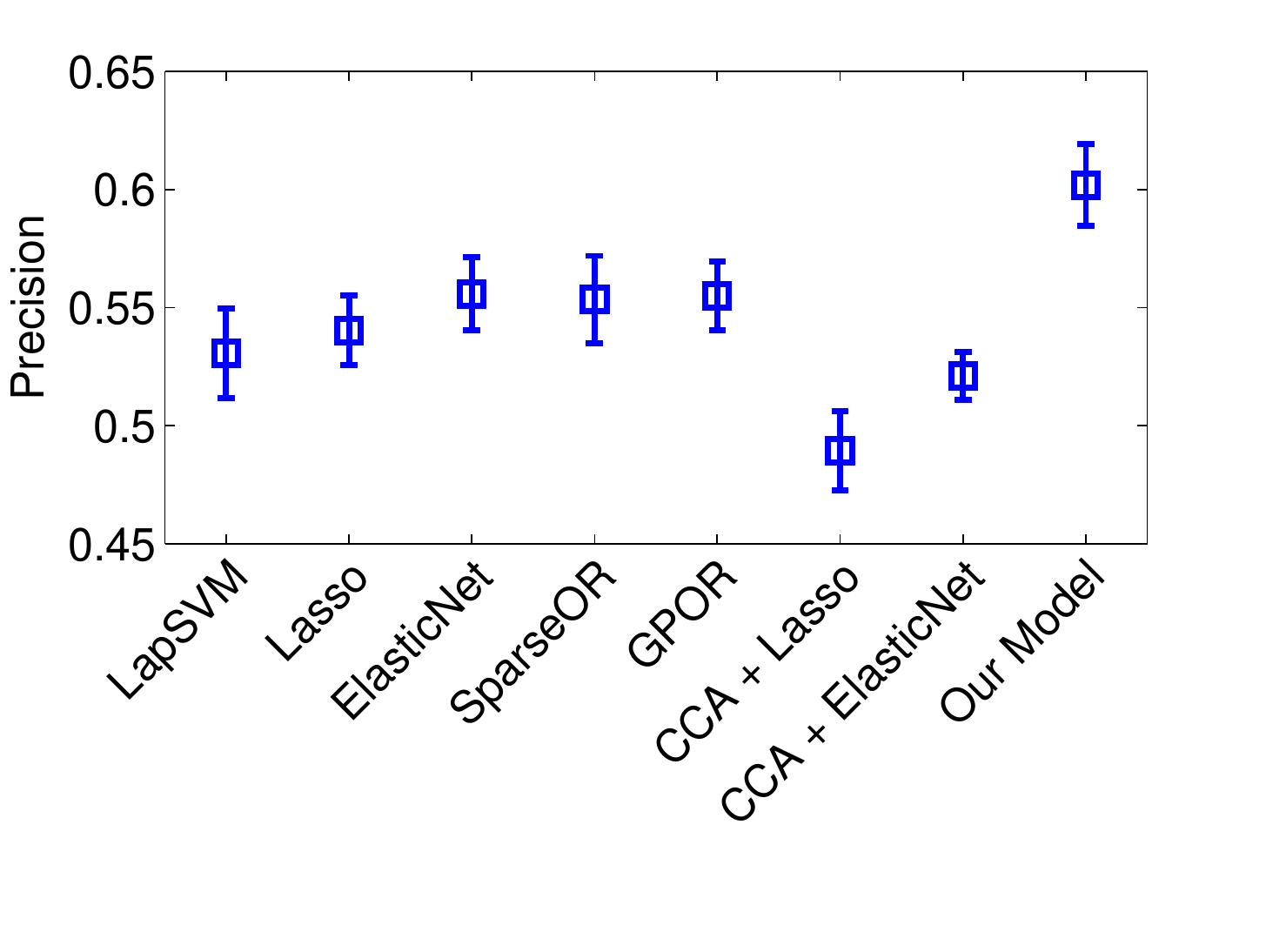}
\vspace{-25pt}
\caption{\small The prediction accuracy with standard errors on the real data.
 }
 \vspace{-10pt}
\label{fig:real_acc}
\end{figure}

The accuracies for predicting unknown labels $\y$ and their standard errors are shown in Figure \ref{fig:real_acc}. Our method achieved the highest prediction accuracy, higher than that of the second best method, GP ordinal Regression, by 10\% and than that of the worst method, CCA+lasso, by 22\%.
\comment{
 We also investigate the prediction performance of the baseline methods only on one single view (SNPs or MRI). We found that compared to using both types of features, using SNPs only got much worse performance and using MRI only got comparable performance. For example, Elastic-net reported the 10-fold cross-validation average precisions were $0.496$ ,$0.561$ and $0.562$
 respectively for SNPs, MRI and both SNPs and MRI. These results illustrates that the genetic variation and brain atrophy provide complementary information for characterizing the disease.
 }

\begin{figure*}[ht]
\centering
\subfigure[]{
\includegraphics[width=0.475\textwidth]{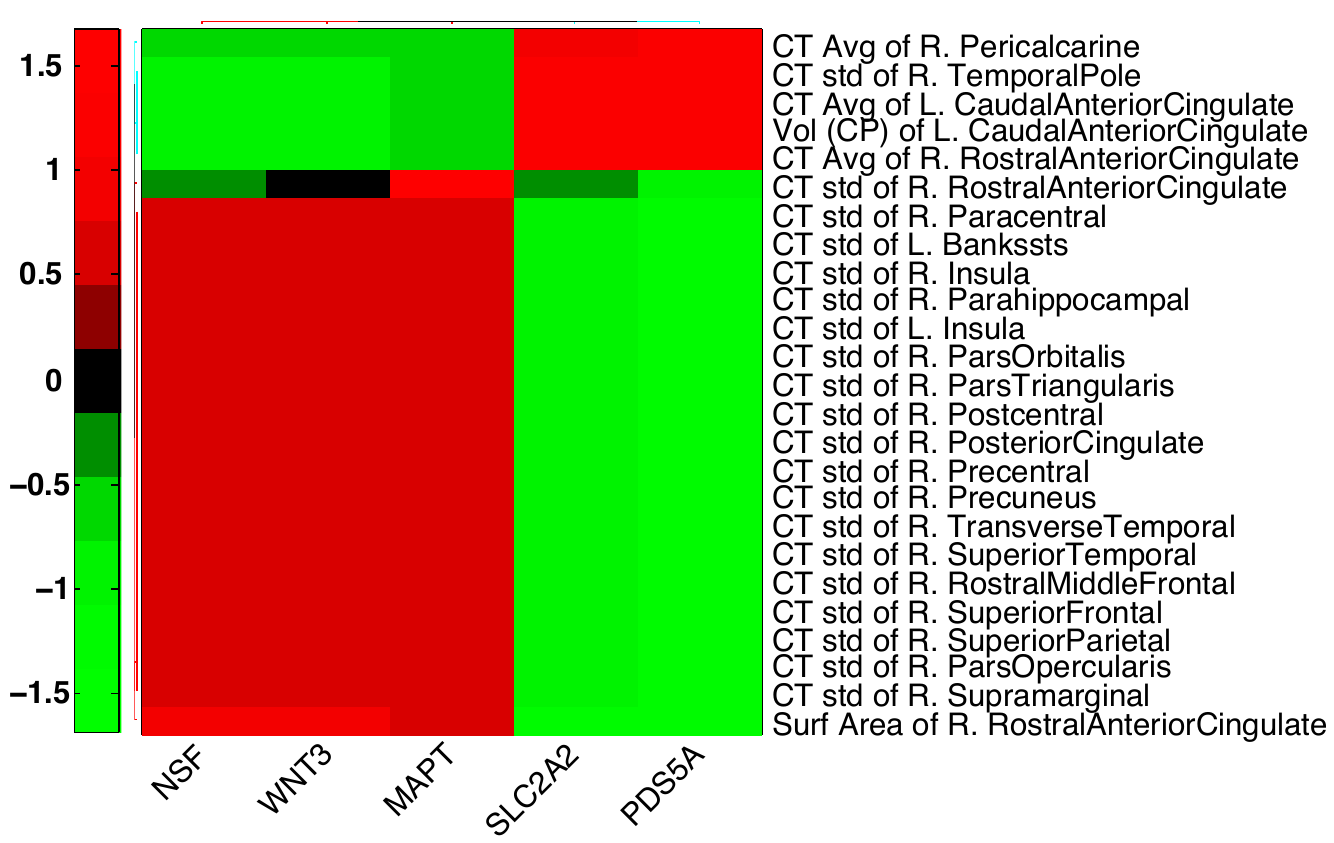}
}
\subfigure[]{
\includegraphics[width=0.475\textwidth]{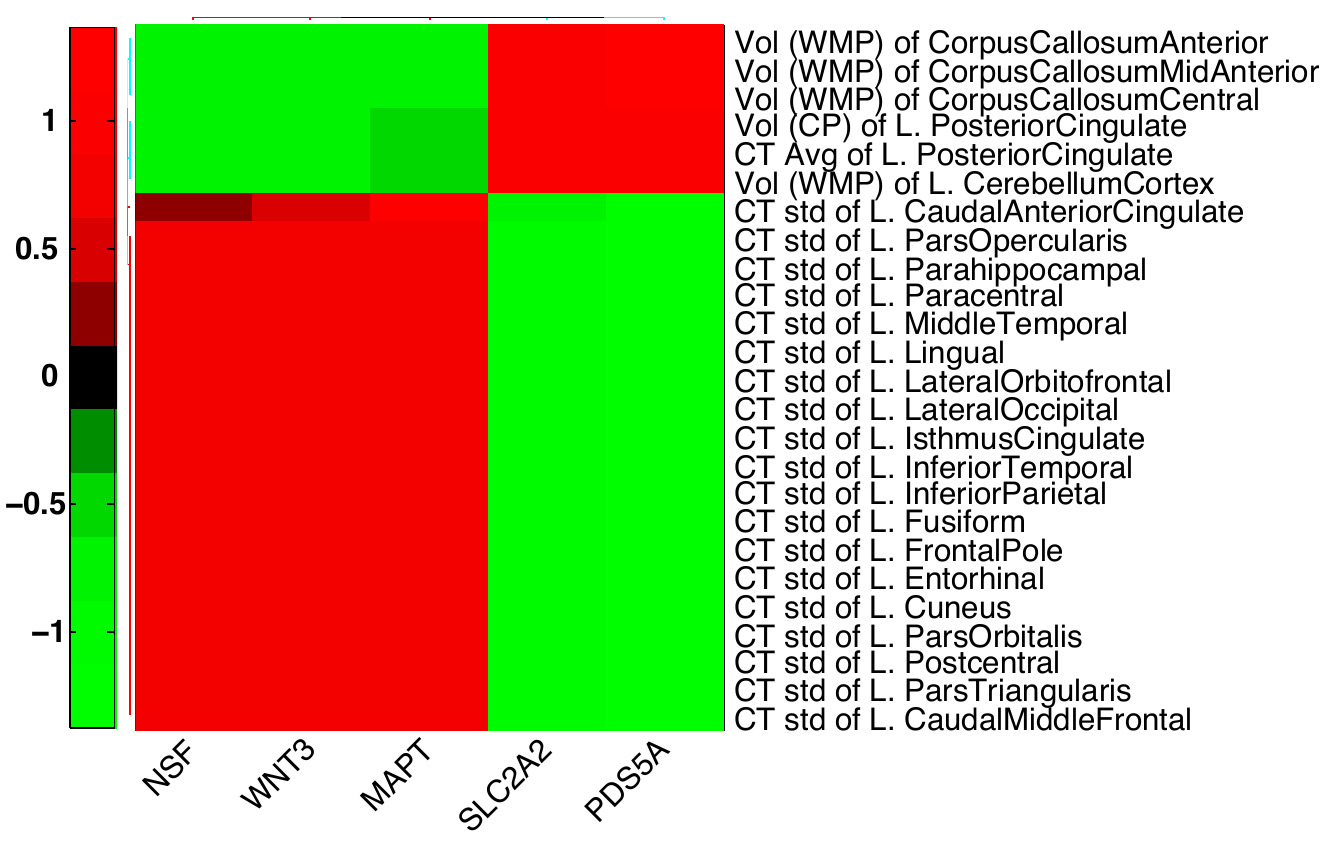}
}
\subfigure[]{
\includegraphics[width=0.475\textwidth]{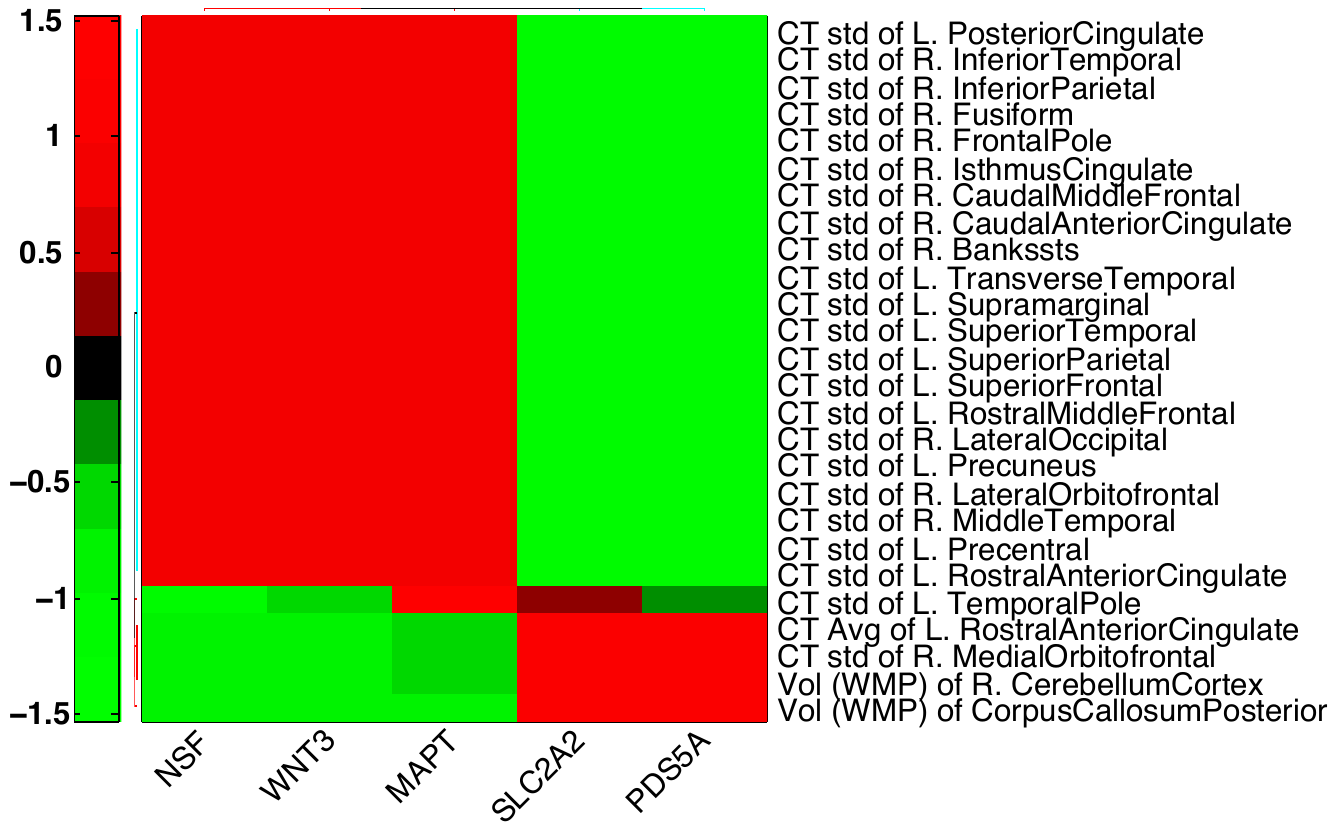}
}
\subfigure[]{
\includegraphics[width=0.475\textwidth]{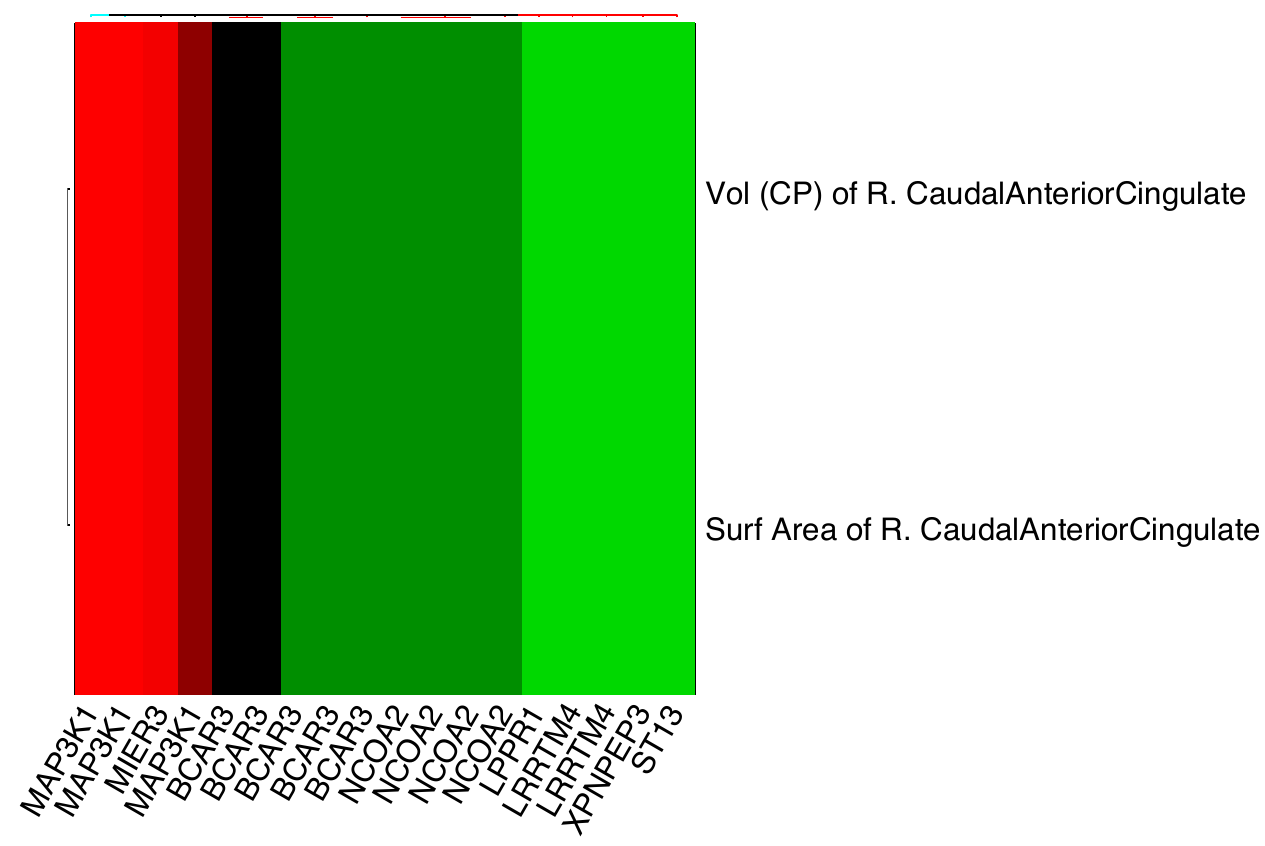}
}
\caption{The estimated associations between MRI features and SNPs. In each sub-figure, the MRI features are listed on the right and the SNP names are given at the bottom.} 
\label{fig:real_linkage}
\end{figure*}


We also examined the strongest associations discovered by \ours based on this dataset. First of all, the ranking of MRI features in terms of their prediction power of three different disease populations (normal, MCI and AD) demonstrate that most of top ranked features are based on the cortical thickness measurement. On the other hand, the features based on volume and surface area estimation of the same brain structures are less predictive. Particularly, thickness measurements of middle temporal lobe, precuneus, and fusiform were found to be most predictive compared with other brain regions. These findings are consistent with the memory-related function in these regions and findings in the literature for their prediction power of AD. We also found that measurements of the same structure on the left and right side have similar weights, indicating that the algorithm can automatically select correlated features in groups, since no asymmetrical relationship has been found for the brain regions involved in AD.

Secondly, the analysis of associating genotype to AD disease prediction also generated interesting results. Similar to the MRI features, SNPs that are in the vicinity of each other often listed together, indicating the group selection characteristics of the algorithm. For example, the top ranks SNPs are associated with a few genes including PSMC1P12 (proteasome 26S subunit, ATPase), NCOA2 (The nuclear receptor coactivator 2), and WDR52(WD repeat domain 52), which have been studied intensively in cancer research.

At last, biclustering of the gene-MRI association, as shown in Figure~\ref{fig:real_linkage} reveal interesting pattern in terms of the relationship between genetic variations and brain atrophy measured by structural MRI. For example, the top ranks SNPs are associated with a few genes including BCAR3 (Breast cancer anti-estrogen resistance protein 3) and NCOA2, which have been studied more carefully in cancer research. One of the genes associated with this set of SNPs is MATP (microtubule-associated protein tau), which codes the tau gene that are associated closely with the AD. These findings reveal strong association between MATP gene and atrophy in the memory-related brain regions. Moreover, the same set of SNPs are also highly associated with cingulate, but in an opposite direction. These results indicate an opposite effect of genotype to the cingulate region, which is part of the limbic system and involve in emotion formation and processing, compared with other structures such as temporal lobe, which plays a more important role in the formation of long-term memory.

In summary, SHML discovered
synergistic predictive relationships between brain atrophy, genetic variations and the disease status.


%

\section{Conclusions}
\label{sec:conclusion}
We have presented, \SHML, a new Bayesian multiview learning framework.
\ours simultaneously finds key associations between data sources (\ie, genetic variations and phenotypic traits) and to predict unknown ordinal labels.
Experimental results on the ADNI data indicate that SHML found biologically meaningful associations between SNPs and MRI features and led to significant improvement on predicting the ordinal AD stages over the alternative classification and ordinal regression methods. Although we have focused on the AD study, we expect that \ours, as a powerful extension of CCA, can be applied to a wide range of applications in biomedical research -- for example, eQTL analysis supervised by additional labeling information.
  %

\section{Acknowledgments}
This work was supported by NSF IIS-0916443, NSF CAREER award IIS-1054903, and the Center for
Science of Information (CSoI), an NSF Science and Technology Center, under grant agreement CCF-0939370.

\bibliographystyle{unsrt}
\begin{small}
\bibliography{sparseLinkOrdinal}  
\end{small}
\end{document}